\pdfoutput=1
\documentclass[11pt]{article}

\usepackage[final]{acl}
\title{From BERT to T5: A Study of Named Entity Recognition}
\author{Mei Jia $^{\dagger}$\thanks{~~This work was completed during the author's Master's studies at the University of Manchester.}\\
  University of Manchester\\
  \texttt{jiamei.meow@gmail.com} \\} 

\usepackage{times}
\usepackage{latexsym}
\usepackage{booktabs}
\usepackage{multirow}
\usepackage[T1]{fontenc}
\usepackage[utf8]{inputenc}
\usepackage{microtype}
\usepackage{inconsolata}
\usepackage{graphicx}
\usepackage{textgreek} 
\usepackage{float}
\usepackage[section]{placeins}
\usepackage{amsmath}
\usepackage{xurl}    
\usepackage{array}
  
\begin{document}
\maketitle
\begin{abstract}
 Named entity recognition (NER) has been one of the essential preliminary steps in modern NLP applications. This report focuses on implementing the NER task on finetuning two pretrained models: (i) an encoder-only model (BERT) with a simple classification head, and (ii) a sequence-to-sequence model (T5) with few-shot prompts. Under the original 7-class tag and 3-class simplified tag schemes, BERT is applied a weighted cross-entropy for training loss, and T5 is fine-tuned with two validation strategies. It also conducted an ablation study with different hyperparameters. Moreover, the related analysis provides valuable insights into common errors in BERT and the two models' performance. Based on a bunch of performance metrics, this report aims to compare the above two architectures and explore their abilities in the sequence labelling task, laying the groundwork for further practical use cases. 
\end{abstract}

\section{Introduction}
\citet{jurafsky2021rule} defined Named Entity Recognition (NER) as identifying spans of text that constitute proper names and tag the related type of entity. Through higher-level semantic and relational analyses, NER converts raw text into structured data, laying the foundations for various applications such as question answering, semantic representation, and so on. Instead of relying on handcrafted rules, the end-to-end deep learning system can learn automatically from massive corpora through pre-trained language models, providing a more efficient way for this work. Therefore, we select an encoder-only model (BERT) and a sequence-to-sequence model (T5) to fine-tune on this classic task.
For the BERT model, its advantages lie in bidirectional context encoding and token-by-token tagging. As shown in related work, this architecture only with a classification head can achieve high performance on different NER datasets (CoNLL03: 93.04\% F-score, OntoNotes5.0: 91.11\% F-score) \citep{li2020survey}. It can serve as a good baseline in this study. Additionally, NER is described as mapping input tokens to a sequence with tags in the "3.4 Tag Decoder Architectures" Section of \citet{li2020survey}'s work. We also consider re-framing the NER task as a text-to-text generation problem through the T5 model. It can be expected to offer greater flexibility in output formats. 
\section{Data}
\begin{figure*}[htbp]
\includegraphics[width=0.5\linewidth]{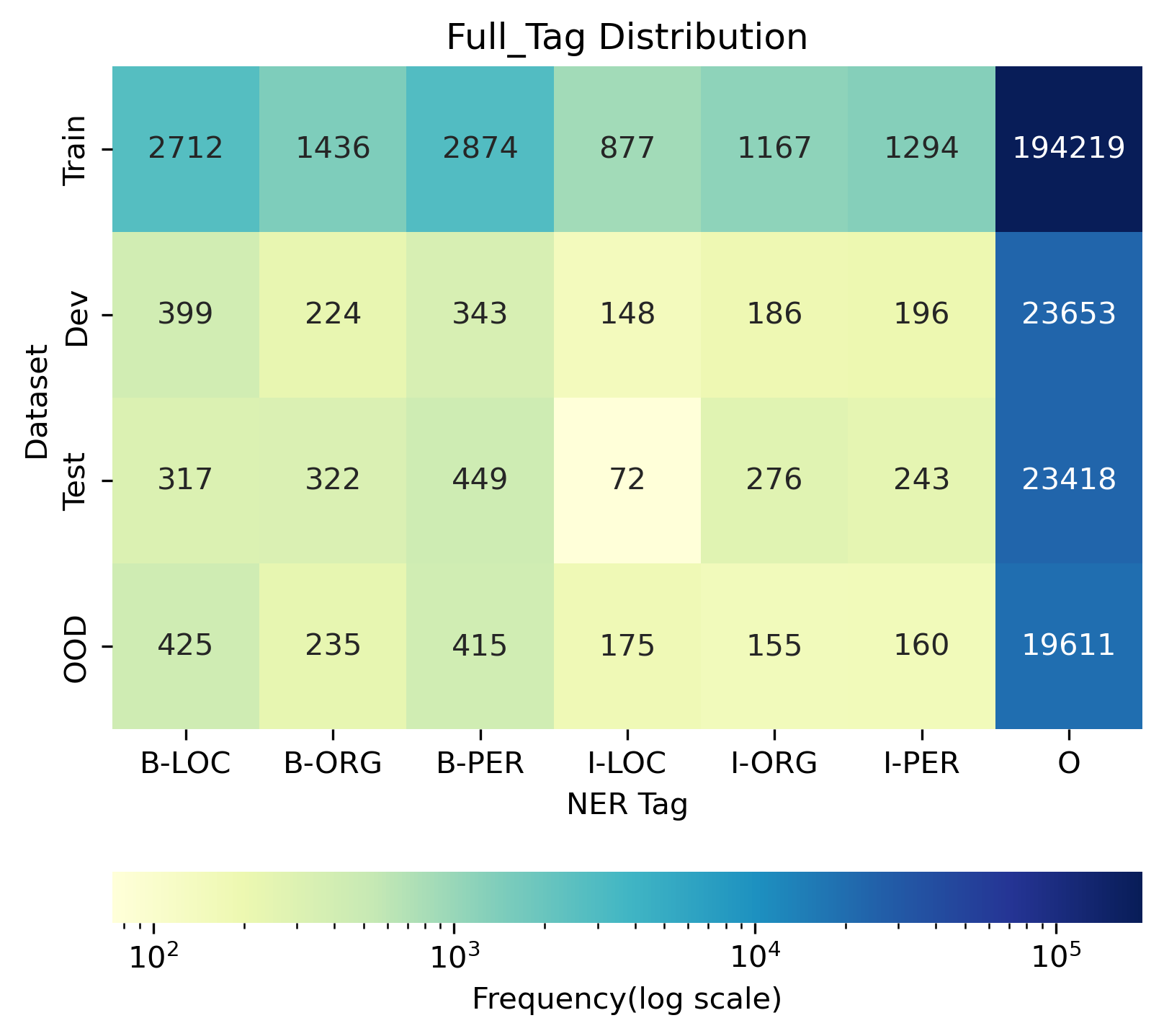} \hfill
  \includegraphics[width=0.32\linewidth]{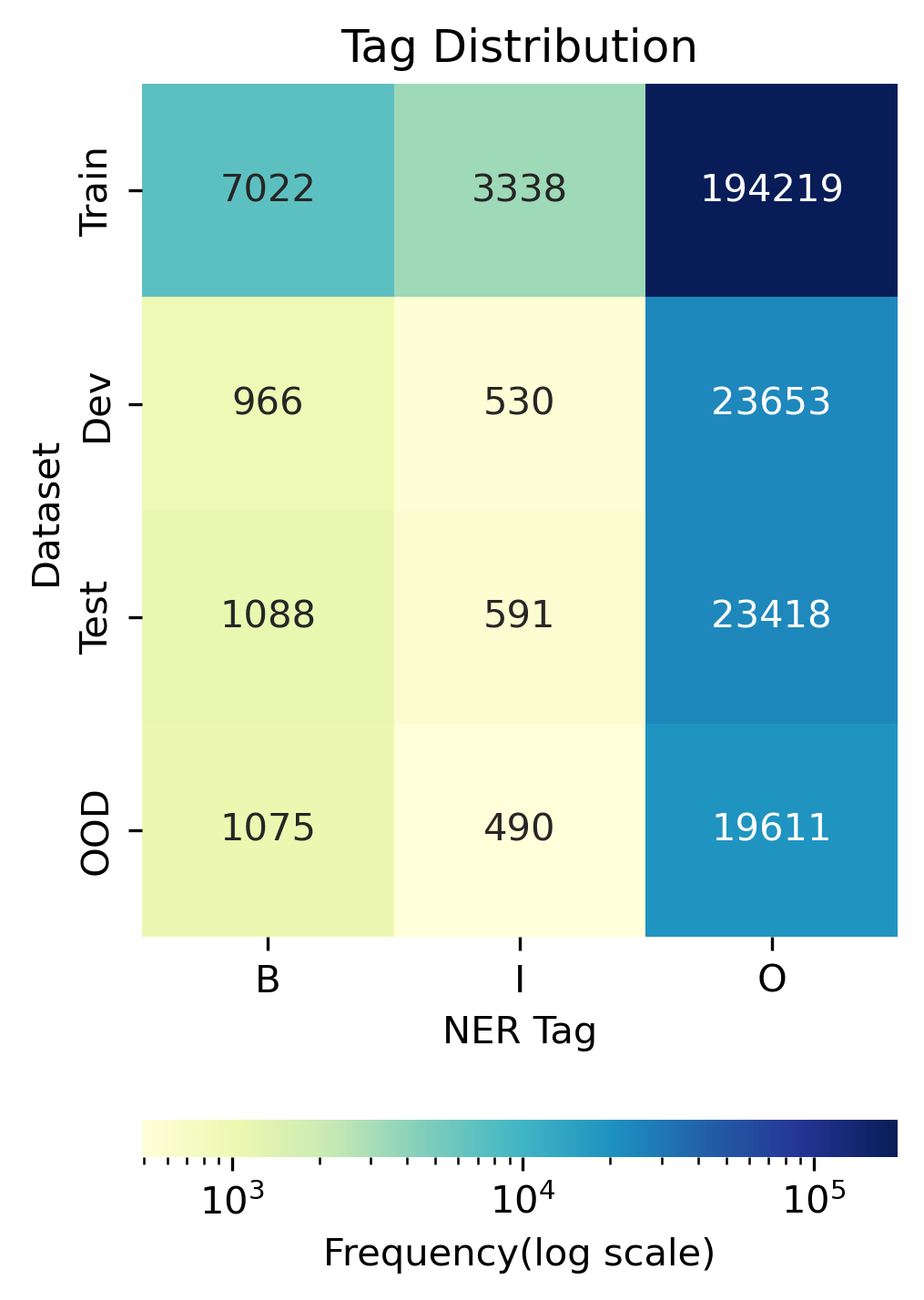}
  \caption {Heatmaps of tag distribution of 7-class tags (left) and 3-class tags (right) across training (Train), validation (Dev), in‑domain test (Test) and out‑of‑domain test (OOD) datasets.}
  \label{fig:entity-dist}
\end{figure*}

This study adopts a preprocessed NER dataset with a similar format of the CoNLL-U scheme, which is split into three primary subsets: training, validation and in-domain test. A second out-of-domain test dataset is also introduced to evaluate the model’s generalisation ability. Two annotation standards are employed: (i) the Full tags, which convert location, person and organisation entity types into six tags (“B-LOC”, “I-LOC”, “B-PER”, “I-PER”, “B-ORG”, “I-ORG”) with an additional non-entity “O” tag, and (ii) the simple tags, which simplify all entity types into more general “BIO” labels (“Beginning”, “Inside” and “Outside”).
As shown in Figure~\ref{fig:entity-dist}, the entity label distribution is heavily skewed toward the “O” tag across all datasets, with non-entity tokens accounting for over 90\%. It highlights the challenge of extracting sparse spans against a wide non-entity context, so we reduce the weights of high-frequency non-entity tags during model training. In the Full tag scheme, an imbalance between beginning (“B-”) and inside (“I-”) entity labels is evident in each dataset. Among three entity types, “PER” spans appear most frequently, followed by “LOC” spans. In the Simple tag scheme, “B” tags exist nearly twice as often as "I" tags, which implies that most non-single-token named entities are only two tokens long. All of these provide essential insights for error analysis.
\section{Experimental Setup}
\subsection{Model}
For the first model, we select a vanilla pre-trained BERT, which is originally trained on Masked Language Modelling and Next Sentence Prediction to learn deep bidirectional representations \citep{devlin2019bert}. To transform contextual embeddings into "BIO" labels, we add a single linear classification head with softmax. The related class-label scores are calculated as follows:
\begin{equation}
  \mathbf{z}_i = \mathbf{W}\,\mathbf{h}_i + \mathbf{b}, 
  \quad
  \label{eq:linear}
\end{equation}
\begin{equation}
  P\bigl(y_i = l \mid \mathbf{h}_i\bigr)
  = \frac{\exp(z_{i,l})}{\sum_{l'} \exp(z_{i,l'})}
  \label{eq:softmax}
\end{equation}
Moreover, we address the extreme imbalance of the non-entity label by using a weighted cross-entropy with a hyperparameter α. We consider several values of α to find the optimal trade-off between reducing false positives on the “O” tag and preserving overall precision. The related formulas are calculated as follows:
\begin{equation}
  \mathcal{L}_{\mathrm{WCE}}
  = -\sum_{i=1}^{n} w_{y_i}\,\log P(y_i \mid x)
  \label{eq:wce}
\end{equation}
\begin{equation}
  w_{l} =
  \begin{cases}
    \alpha, & l = "O",\\
    1,      & \text{otherwise},
  \end{cases}
  \quad \alpha \in \{0.7,\,0.8,\,1\}.
  \label{eq:weights}
\end{equation}
The second model is a vanilla pre-trained T5 \citep{raffel2020t5}, which learns to generate outputs from partial inputs through an encoder-decoder architecture. During NER fine-tuning, the encoder processes each input sentence and the decoder is trained to autoregressively generate the corresponding output sequence of "BIO" labels by maximising the joint conditional likelihood. The related formulas are calculated as follows:
\begin{equation}
  \mathbf{Y} = \arg\max_{(y_{1},\dots,y_{n})} P(Y \mid X)
  \label{eq:decode}
\end{equation}
\begin{equation}
  P(Y \mid X)
  = \prod_{t=1}^{n} P\bigl(y_{t} \mid y_{<t},\,X\bigr).
  \label{eq:factorization}
\end{equation}

\subsection{Fine-tuning Details}
\begin{table*}[htbp]
  \centering
  \small
  \renewcommand{\arraystretch}{1.2}
  \begin{tabular}{ 
    >{\centering\arraybackslash}p{1cm} 
    p{9cm} }
    \toprule
    \textbf{Number}
      & \textbf{Prompt Selection} \\
    
    \hline
        \addlinespace[0.5ex]
    1  & "Sentence: (word string) [SEP] label each token with its entity type:" \\
    2  & "ner:" + (word string) \\
    3  & "Sentence: (word string) [\textbackslash n] Label tokens:" \\
    4 & "Sentence:[\textbackslash n]" + (word string) + [\textbackslash n] + "Tags:" \\
    \bottomrule
  \end{tabular}
  \caption{Different prompt selections during T5 fine-tuning.}
  \label{tab:prompts}
  \end{table*}

For each model, we selectively adjusted different optimisation-related hyperparameters. We applied 20 epochs and early stopping with 5 epochs for every configuration, and retained the best model checkpoint with the highest score, as shown in Figure~\ref{tab:BERT_Full}-~\ref{tab:T5_Simple} in the Appendix. Ultimately, we used AdamW optimiser and selected a configuration with a batch size of 32 and a learning rate of 1e-5 for BERT and 1e-4 for T5.

For BERT training, we excluded the [CLS] and [SEP] tokens. We also considered a weighted cross-entropy with α = 0.8. For T5 training, we tried several prompts as shown in the Table~\ref{tab:prompts} and selected the last one with a few-shot scheme with two examples as below:\\
\hspace*{1.5em}Sentence: \texttt{"Dmitry went to Moscow ."}\\
\hspace*{1.5em}Tag: \texttt{"B-PER O O B-LOC O"}\\
\hspace*{1.5em}Sentence: \texttt{"Colin has a dog ."}\\
\hspace*{1.5em}Tag: \texttt{"B-PER O O O O"}\\

Although prompt engineering is proven to be critical to T5, the change in single-sentence templates contributes slightly in my experiments, while the few-shot scheme with fine-tuning directly improves the model’s performance by adding only two examples.

Unlike token-level classification, sequence labelling tasks require models to both detect the correct boundaries of each entity and assign the related class. The learning objective of models is to minimise the error between the predicted tag and span and the gold annotation. Being biased by the highly skewed non-entity tags, standard token-level accuracy encourages the models to achieve spuriously high scores by predicting the “O” tag for every token. It will yield seemingly strong performance without any actual learning in early training. Therefore, we considered the labelled span-level F1 as the metric of early stopping checkpoints on the validation dataset, with related precision and recall monitored at the same time, to reflect a more reliable performance indicator. 
  
For T5 validation, we compared two strategies. In my initial few experiments, we tried to directly extract the model’s logits and likelihoods due to training efficiency. However, it tended to misunderstand and generate incorrect entity tags, leading to high recall and low precision as shown in the Table~\ref{tab:T5_valstr}. We decided to go ahead with the autoregressive “model.generate” procedure (num\_beams=3) as it outperformed the likelihood-based shortcut in the majority of experiments by imposing a global constraint over the entire sequence. 

For overall evaluation in the Test and OOD dataset, we used two kinds of performance metrics for both the 7-class and 3-class tag schemes: (i) labelled and unlabelled span matching score; (ii) labelled span-level precision, recall and F1 as well as the related macro scores. Based on different tag types, we also evaluated the full tag scheme with separate precision, recall and F1 to complement the experimental results. Related source code is available at:\url{https://anonymous.4open.science/r/LELA60332_From-BERT-to-T5-A-Study-of-Named-Entity-Recognition-4002/}.
\begin{table}[t]
  \centering
  \small
   \renewcommand{\arraystretch}{1.2}
  \begin{tabular}{ 
    >{\centering\arraybackslash}p{2.5cm} 
    >{\centering\arraybackslash}p{1.5cm} >{\centering\arraybackslash}p{1.5cm}
  }
    \toprule
    \multirow{2}{*}[-1ex]{\bfseries Metric (T5)}
      & \multicolumn{2}{c}{\bfseries Validation Strategy}\\
    \cmidrule(lr){2-3}

      &{\bfseries Likelihood}
      & {\bfseries Generate} \\
    \midrule
    \addlinespace[0.5ex]
    \multicolumn{3}{c}{\bfseries Full Tag} \\
    \addlinespace[0.5ex]
    \hline
    Best-model Epoch  & 12 & 7\\
    \hline
    \addlinespace[0.5ex]
    Precision  & 10.5 & 64.4\\
    Recall  & 79.5 & 65.3\\
    F1 Score  & 18.5 & 64.9\\
    \hline
    \addlinespace[0.5ex]
    \multicolumn{3}{c}{\bfseries Simple Tag} \\
    \addlinespace[0.5ex]
    \hline
    Best-model Epoch  & - & 8\\
    \hline
    \addlinespace[0.5ex]
    Precision  & - & 67.9\\
    Recall  & - & 71.3 \\
    F1 Score  & - & 69.6 \\
\bottomrule
  \end{tabular}
  \caption{Labelled-span-level Performance of T5 model with two strategies under learning rate (1e-4), for full and simple tag schemes in validation datasets.}
 \label{tab:T5_valstr}
  \end{table}

\section{Result}
\begin{table*}[t]
  \centering
  \small
  \renewcommand{\arraystretch}{1.2}

  \begin{tabular}{ 
    >{\centering\arraybackslash}p{2.5cm} 
    >{\centering\arraybackslash}p{1.5cm} >{\centering\arraybackslash}p{1.5cm} >{\centering\arraybackslash}p{1.5cm} >{\centering\arraybackslash}p{1.5cm}
  }
    \toprule
    \multirow{2}{*}[-1ex]{\bfseries Metric (BERT)}
      & \multicolumn{2}{c}{\bfseries Full Tag}
      & \multicolumn{2}{c}{\bfseries Simple Tag} \\
    \cmidrule(lr){2-3} \cmidrule(lr){4-5}

      & {\bfseries Test}
      & {\bfseries OOD}
      & {\bfseries Test}
      & {\bfseries OOD} \\
    \addlinespace[0.5ex]
    \midrule
    \addlinespace[0.5ex]
    \multicolumn{5}{c}{\bfseries Labelled Span} \\
    \addlinespace[0.5ex]
    \hline
    Matching Score  & 84.8 & 72.8 & 87.6 & 79.7 \\
    Precision  & 84.6 & 80.7 & 86.4 & 85.0 \\
    Recall  & 84.8 & 72.8 & 87.6 & 79.7 \\
    F1 Score  & 84.7 & 76.6 & 87.0 & 82.3 \\
    Macro Precision  & 81.4 & 77.4 & 85.2 & 84.2 \\
    Macro Recall  & 83.6 & 68.4 & 87.8 & 80.6 \\
    Macro F1 Score  & 82.4 & 71.3 & 86.5 & 82.4 \\
    Accuracy  & 98.3 & 97.8 & 98.5 & 98.2 \\
    \hline
    \addlinespace[0.5ex]
    \multicolumn{5}{c}{\bfseries Unlabelled Span} \\
    \addlinespace[0.5ex]
    \hline
    Matching Score   & 87.7 & 79.3 & 87.6 & 79.7 \\
    \bottomrule
  \end{tabular}
  \caption{Performance of BERT model with full and simple tag schemes on both labelled‑span and unlabelled‑span metrics across in‑domain (Test) and out‑of‑domain (OOD) datasets.}
  \label{tab:BERT_Performance}
  \end{table*}

\begin{table*}[t]
  \centering
  \small
  \renewcommand{\arraystretch}{1.2}

  \begin{tabular}{ 
    >{\centering\arraybackslash}p{2.5cm} 
    >{\centering\arraybackslash}p{1.5cm} >{\centering\arraybackslash}p{1.5cm} >{\centering\arraybackslash}p{1.5cm} >{\centering\arraybackslash}p{1.5cm}
  }
    \toprule
    \multirow{2}{*}[-1ex]{\bfseries Metric (T5)}
      & \multicolumn{2}{c}{\bfseries Full Tag}
      & \multicolumn{2}{c}{\bfseries Simple Tag} \\
    \cmidrule(lr){2-3} \cmidrule(lr){4-5}

      & {\bfseries Test}
      & {\bfseries OOD}
      & {\bfseries Test}
      & {\bfseries OOD} \\
    \addlinespace[0.5ex]
    \midrule
    \addlinespace[0.5ex]
    \multicolumn{5}{c}{\bfseries Labelled Span} \\
    \addlinespace[0.5ex]
    \hline
    Matching Score  & 68.6 & 57.2 & 73.7 & 64.9 \\
    Precision  & 67.6 & 60.8 & 70.0 & 66.8 \\
    Recall  & 68.6 & 57.2 & 73.7 & 64.9 \\
    F1 Score  & 68.1 & 58.9 & 71.8 & 65.8 \\
    Macro Precision  & 65.7 & 61.4 & 70.6 & 67.1 \\
    Macro Recall  & 66.6 & 54.1 & 73.7 & 64.9 \\
    Macro F1 Score  & 66.1 & 56.6 & 72.1 & 66.0 \\
    \hline
    \addlinespace[0.5ex]
    \multicolumn{5}{c}{\bfseries Unlabelled Span} \\
    \addlinespace[0.5ex]
    \hline
    Matching Score   & 72.1 & 63.6 & 73.7 & 64.9 \\
    \bottomrule
  \end{tabular}
  \caption{Performance of T5 model with full and simple tag schemes on both labelled‑span and unlabelled‑span metrics across in‑domain (Test) and out‑of‑domain (OOD) datasets.}
  \label{tab:T5_Performance}
  \end{table*}

\begin{table*}[htbp]
  \centering
  \small
  \renewcommand{\arraystretch}{1.2}

\begin{tabular}{
    >{\centering\arraybackslash}p{2cm} 
    >{\centering\arraybackslash}p{1cm} >{\centering\arraybackslash}p{1cm} 
    >{\centering\arraybackslash}p{1cm} >{\centering\arraybackslash}p{1cm} 
    >{\centering\arraybackslash}p{1cm} >{\centering\arraybackslash}p{1cm}
  }
  \toprule
  \multirow{2}{*}[-0.5ex]{\bfseries Metric (BERT)}
    & \multicolumn{2}{c}{\bfseries "LOC"}
    & \multicolumn{2}{c}{\bfseries "ORG"}
    & \multicolumn{2}{c}{\bfseries "PER"} \\
  \cmidrule(lr){2-3} \cmidrule(lr){4-5} \cmidrule(lr){6-7}

    & {\bfseries Test} & {\bfseries OOD}
    & {\bfseries Test} & {\bfseries OOD} 
    & {\bfseries Test} & {\bfseries OOD} \\
  \midrule
  Precision  & 79.8 & 70.8 & 72.1 & 72.5 & 92.2 & 88.9 \\
  Recall     & 87.0 & 77.4 & 69.0 & 40.4 & 94.9 & 87.2 \\
  F1 Score   & 83.3 & 73.9 & 70.5 & 51.9 & 93.5 & 88.1 \\
  \bottomrule
\end{tabular}
  \caption{Labelled-span-level Performance of BERT model with full tag schemes across in‑domain (Test) and out‑of‑domain (OOD) datasets, for different tag types.}
  \label{tab:BERTtag_Performance}
  \end{table*}
  
\begin{table*}[htbp]
  \centering
  \small
  \renewcommand{\arraystretch}{1.2}  
\begin{tabular}{ 
    >{\centering\arraybackslash}p{2cm} 
    >{\centering\arraybackslash}p{1cm} >{\centering\arraybackslash}p{1cm} >{\centering\arraybackslash}p{1cm} >{\centering\arraybackslash}p{1cm} >{\centering\arraybackslash}p{1cm} >{\centering\arraybackslash}p{1cm}
  }
    \toprule
    \multirow{2}{*}[-1ex]{\bfseries Metric(T5)}
      & \multicolumn{2}{c}{\bfseries "LOC"}
      & \multicolumn{2}{c}{\bfseries "ORG"}
      & \multicolumn{2}{c}{\bfseries "PER"} \\
    \cmidrule(lr){2-3} \cmidrule(lr){4-5} \cmidrule(lr){6-7}

      & {\bfseries Test}
      & {\bfseries OOD}
      & {\bfseries Test}
      & {\bfseries OOD} 
      & {\bfseries Test}
      & {\bfseries OOD}\\
    \addlinespace[0.5ex]
    \hline
    Precision  & 64.6 & 53.9 & 53.1 & 62.5 & 79.5 & 67.8 \\
    Recall  & 65.0 & 56.7 & 50.9 & 36.2 & 83.8 & 69.6 \\
    F1 Score  & 64.8 & 55.2 & 52.0 & 45.8 & 81.6 & 68.7 \\
    \bottomrule
  \end{tabular}
  \caption{Labelled-span-level Performance of T5 model with full tag schemes across in‑domain (Test) and out‑of‑domain (OOD) datasets, for different tag types.}
  \label{tab:T5tag_Performance}
  \end{table*}
  
\subsection{Overall Performance}
The detailed performance comparison of the above two models with both Full tag and Simple tag differences listed in Table~\ref{tab:BERT_Performance} and ~\ref{tab:BERTtag_Performance} for BERT and Table~\ref{tab:T5_Performance} and ~\ref{tab:T5tag_Performance} for T5, which displays the shared behaviours and difference between these two models. Under the level of labelled span, BERT achieves an 84.7\% F1 score with the Full tag scheme in the test dataset and related macro F1 of 82.4\%. When it comes to simple tags, it improves to 87.0\% and 86.5\%. A similar situation happens in T5, the metric results jumping by 3-6 percentage points respectively. The Matching score also increases dramatically when changing to the case of an unlabelled span. In contrast, the OOD dataset brings more challenges to both models with the same decreased trend. For example, T5 with full tags dropped about 8-12 points among all metrics. 

On one hand, both BERT and T5 show several consistent trends. First, they benefit from the simpler "BIO" scheme and the less stringent unlabelled span matching score. Compared to the full tag scheme, a modest improvement appears in the simple case of both models, showing that more remaining errors are related to type confusions rather than span mismatching. Second, when evaluated on the OOD dataset with longer samples and low-frequency tokens, they both suffer a direct performance drop. More interestingly, BERT and T5 both show the recall-dominant behaviour in the Test dataset and the precision-dominant action in the OOD dataset. When meeting unfamiliar cases, the models tend to predict the safer “O” tags and avoid other entities, leading to a reduction of false positives but missing more true spans. This suggests more space for improving the generalisation ability of these two models. Third, macro F1 remains below their F1 scores, which reveals the imbalanced effects across different entity tags and the performance is dominated by more frequent classes. Finally, the models follow the same ordering of recognition difficulty as “PER” easiest, “LOC” intermediate and “ORG” hardest. On the other hand, the two models have significant differences. First, BERT strongly outperforms T5 under every metric and domain, especially in the OOD dataset. While the OOD performance of T5 falls sharply regardless of the tag schemes, BERT decreases in a reasonable range when converting the domain. These findings indicate the greater robustness of fine-tuning encoder-only architectures to the sequence labelling task. Second, a simple tag scheme yields greater gains for T5 (e.g.,  In-domin F1 score: +10.9\%, Out-of-domain Labelled matching score: +7.7\%), which highlights the sequence-to-sequence model struggling more with identifying entity types. Third, BERT shows a larger precision-recall imbalance when shifting into the OOD dataset (e.g., F1 score with full tags: -7.9\%, F1 score with simple tags: -5.3\%), approximately twice the gap observed for T5 under two tag schemes. This reflects that BERT is more sensitive to the data distribution due to its token-wise labelling strategy, which is more likely to assign the non-entity tags to every position when the model feels less confident in the OOD data. In contrast, the decoding strategy of T5 generates the entire tag sequence in one pass and has a smaller precision-recall gap.

Overall, an encoder-only architecture provides an ideal solution for span-level NER, and the sequence-to-sequence model needs further optimisation to bridge the performance gap.
\subsection{Hyperparameter Sensitivity}
The change of hyperparameters indeed has effects on the performance of both models. First, we explored different learning rates to find the optimal configurations for BERT (Table~\ref{tab:BERT_lr}) and T5 (1e-4) based on labelled span-level F1. A higher rate causes under-fitting, while a lower rate slows convergence speed, which both degrade most metric scores.

Second, as shown in Table~\ref{tab:BERT_WCE}, a weighted cross-entropy with $\alpha$ = 0.8 leads to the highest validation F1 (82.4\%) with a precision-recall balance in BERT. The unweighted baseline ($\alpha$ = 1) and the other weighted value ($\alpha$ = 0.7) both drop slightly at 82.1\%, while the latter has more balanced results with precision (+1.6\%) and recall (-1.6\%)\footnote{A more aggressive discount like “$\alpha$ = 0.2” leads to a worse score, so we not record in the table.}. These findings indicates that a moderately weighted adjustment of non-entity tags helps the model perform better. However, excessive suppression of the “O” tag harms the results, as the model will predict more nonexistent entities and increase the cases of false positives.

\begin{table}[htbp]
  \centering
  \small
   \renewcommand{\arraystretch}{1.2}
  \begin{tabular}{ 
    >{\centering\arraybackslash}p{2.5cm} 
    >{\centering\arraybackslash}p{0.7cm} >{\centering\arraybackslash}p{0.7cm} >{\centering\arraybackslash}p{0.7cm} >{\centering\arraybackslash}p{0.7cm}
  }
    \toprule
    \multirow{2}{*}[-1ex]{\bfseries Metric (BERT)}
      & \multicolumn{4}{c}{\bfseries Learning Rate}\\
    \cmidrule(lr){2-5}

      & {\bfseries 1e-4}
      & {\bfseries 2e-5}
      & {\bfseries 1e-5}
      & {\bfseries 1e-6} \\
          \hline
      Best-model Epoch  & 4 & 16 & 20 & 9 \\
    \midrule
    Precision  & 72.7 & 81.6 & 80.4 & 75.1 \\
    Recall  & 75.7 & 80.7 & 83.9 & 80.7 \\
    F1 Score  & 74.2 & 81.2 & 82.1 & 77.8 \\
    Accuracy  & 97.8 & 98.4 & 98.3 & 98.1 \\
    \bottomrule
  \end{tabular}
  \caption{Labelled-span-level Performance of BERT model with different learning rates under weight rate $\alpha$, for full tag schemes in validation datasets.}
  \label{tab:BERT_lr}
  \end{table}
\begin{table}[htbp]
  \centering
  \small
   \renewcommand{\arraystretch}{1.2}
  \begin{tabular}{ 
    >{\centering\arraybackslash}p{2.5cm} 
    >{\centering\arraybackslash}p{1cm} >{\centering\arraybackslash}p{1cm} >{\centering\arraybackslash}p{1cm}
  }
    \toprule
    \multirow{2}{*}[-1ex]{\bfseries Metric (BERT)}
      & \multicolumn{3}{c}{\bfseries Weight Rate $\alpha$}\\
    \cmidrule(lr){2-4}

      &{\bfseries 0.7}
      & {\bfseries 0.8}
      & {\bfseries 1} \\
    \midrule
    \addlinespace[0.5ex]
    \multicolumn{4}{c}{\bfseries Full Tag} \\
    \addlinespace[0.5ex]
    \hline
    Best-model Epoch  & 19 & 18 & 16 \\
    \hline
    \addlinespace[0.5ex]
    Precision  & 82.0 & 82.0 & 80.4 \\
    Recall  & 82.3 & 82.9 & 83.9  \\
    F1 Score  & 82.1 & 82.4 & 82.1  \\
    Accuracy  & 98.4 & 98.4 & 98.3 \\
    \hline
    \addlinespace[0.5ex]
    \multicolumn{4}{c}{\bfseries Simple Tag} \\
    \addlinespace[0.5ex]
    \hline
    Best-model Epoch  & - & 17 & 18 \\
    \hline
    \addlinespace[0.5ex]
    Precision  & - & 85.5 & 86.9 \\
    Recall  & - & 88.1 & 85.3  \\
    F1 Score  & - & 86.8 & 86.1  \\
    Accuracy  & - & 98.8 & 98.8 \\
\bottomrule
  \end{tabular}
  \caption{Labelled-span-level Performance of BERT with different weight rates $\alpha$ under learning rate (1e-5), for full and simple tag schemes in validation set.}
  \label{tab:BERT_WCE}
  \end{table}
  
\subsection{Error Analysis}
In the Table~\ref{tab:BERTtag_Performance} and Table~\ref{tab:T5tag_Performance}, “ORG” is the hardest tag to recognise. This is also consistent with the phenomenon in Figure~\ref{fig:entity-dist} that the "ORG" tag appears the least in the distribution across all datasets. It provides fewer examples for both BERT and T5 models to learn, leading to their confusion about this class type. For example, in the OOD dataset with Full tag in Appendix Figure~\ref{tag:error_Full_OOD} (No.0, row 1), BERT wrongly labels the staff title “Obama special assistant” as “B-PER”. The model also misunderstands “ORG” with “LOC” when multiple entities “burger king portfolios white castle” appear together in the Appendix Figure~\ref{tag:error_Full_Test} (No.112, row 8). These reveal that the models tend to associate some unfamiliar “ORG” entities with other frequent tags. 

Additionally, boundary errors are more likely to exist in longer sentences with more terminology. In Appendix Figure~\ref{tag:error_Full_OOD} (No.8, row 3), BERT only makes a prediction of the first entity “Washington” and omits the rest of the spans like “Metropolitan Club”. Even when converting to the simple tag scheme, the same error happens in the same places (Appendix Figure~\ref{tag:error_Simple_OOD}, (No.8, row 3)), which shows the model struggles with processing multiple spans appearing in the same sentence, especially when meeting low-frequency or unknown entities. More plots about BERT for top error examples across datasets can be viewed in Appendix Figure~\ref{tag:error_Full_Test} and Figure~\ref{tag:error_Full_OOD} for full tags, Figure~\ref{tag:error_Simple_Test} and Figure~\ref{tag:error_Simple_OOD} for simple tags.

The errors on the NER task are partly due to the mismatch between token-level training and span-level evaluation, as mentioned in \citet[Sec.~17]{jurafsky2021rule}. The loss is computed on individual "BIO" tokens, whereas evaluation needs an exact span match. Although span-level F1 is a more appropriate measure for the final output quality, using it for early stopping in a token-level classification work may still cause a degree of validation error. Any boundary misalignment directly affects the final F1 score, and this becomes even worse when models meet OOD datasets.
 
\section{Discussion}
\subsection{Performance Analysis}
In the experimental results, BERT and T5 models can both complete the NER task to some extent. However, BERT consistently outperforms T5 across all evaluation settings, with the performance gap widening most significantly in the OOD dataset. This needs a closer investigation into their performance. 

On one hand, the Transformer-based models give each token a wider range of context and benefits for the disambiguation of entity boundaries and types. When segmenting the spans, the self-attention mechanism is excellent at capturing dependencies from local neighbouring tokens to global context over long distances. It can not only extract spans more precisely by integrating information from both preceding and following tokens, but can also distinguish different meanings of the same word under different cases in the type classification task.

On the other hand, BERT’s strength on the NER task is from its token-level masked language modelling pretraining. As an encoder-only model, it can simultaneously calculate the entire input sequence and make decisions in parallel, providing more efficient and accurate token-wise classification. In contrast, T5 has to generate tokens one by one in time steps, resulting in lower decoding efficiency. Its autoregressive generation also means that an incorrect label on the previous token can accumulate errors and skew all subsequent labels. All of these lead to its disadvantage in the entity tagging task. Additionally, T5 may generate irregular predictions or be prematurely truncated by the termination symbol “<s/>”. This aspect makes T5 perform even worse in the OOD dataset when processing longer inputs or sentences with dense information.

\subsection{Future Direction}
To strengthen the robustness and generalization ability of the above two models, further research will explore two specific directions. First, for the BERT-based architecture, it would be a better choice to integrate a Conditional Random Field (CRF) as the output layer \citep{li2020survey}. With Viterbi decoding, the CRF layer can impose models to focus more on valid "BIO" tags and thus improve the span boundary consistency. Second, for the T5-based model, the experimental result supports a future direction to let the models benefit more from learning the few-shot examples. More diverse prompts can be considered during training and inference stages, particularly driving the model to be exposed to unfamiliar contexts, such as longer sequences with multiple spans or containing more low-frequency entities. 

\section{Conclusion}
This report provides a controlled comparison between an encoder-only model (BERT) and a sequence-to-sequence model (T5) on the NER task. BERT achieves better performance across domains. However, it suffers more from the imbalance between precision and recall. It struggles to process unfamiliar organisation entities or sentences with multiple entities, especially when meeting several closely adjacent spans in the same sentence. Despite being more flexible, the T5 model is more prone to mislabeling and faces low efficiency. Overall, BERT remains the more reliable option for the NER task in practice. Future work could consider more architectural improvements for BERT and diverse few-shot learning for T5.

\bibliography{custom}

\clearpage
\onecolumn
\appendix
\section{Related Figures}
\subsection{Loss Trends and Metrics}
\hspace{30pt}
\begin{figure*}[htbp]
\includegraphics[width=0.5\linewidth]{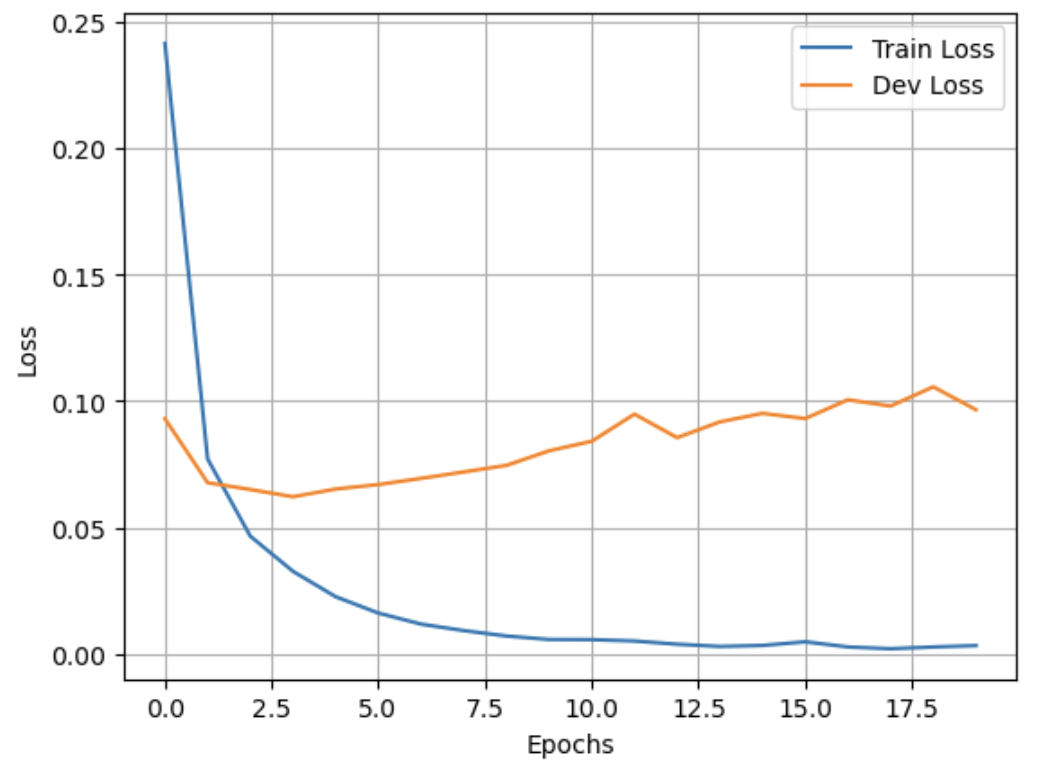} \hfill
\includegraphics[width=0.5\linewidth]{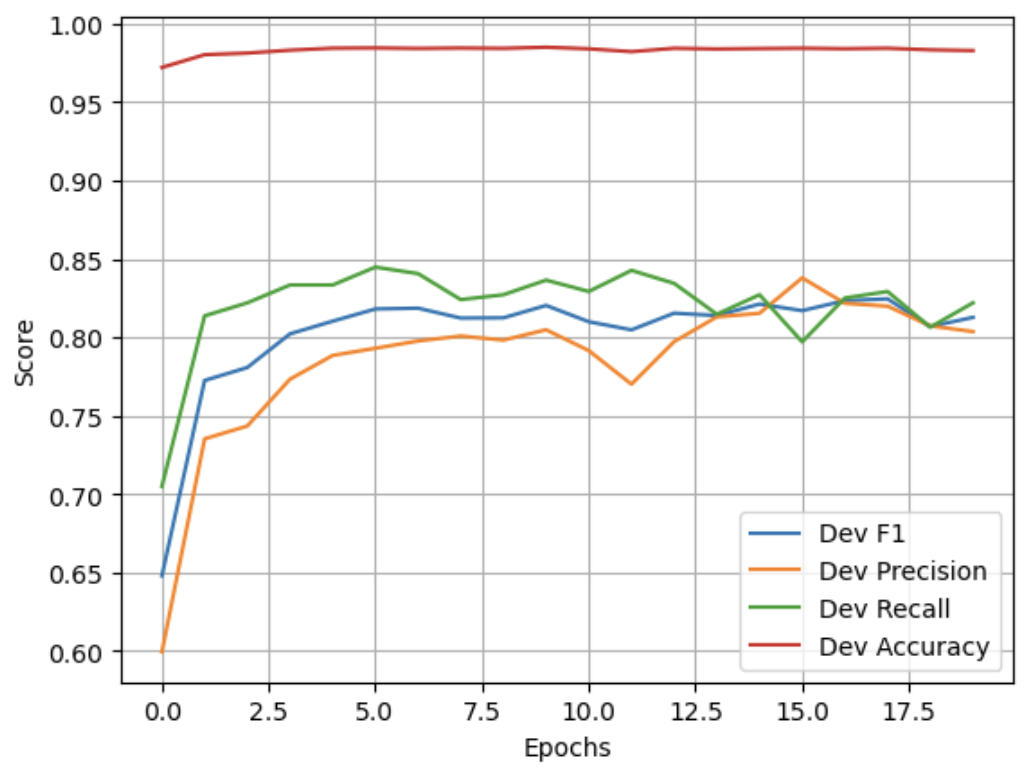}
\caption {Plots of training and validation loss (left) and validation metrics(right) over 20 epochs for the performance of BERT on the full‐tag NER task.}
\label{tab:BERT_Full}
\end{figure*}
\hspace{70pt}
\begin{figure*}[htbp]
\includegraphics[width=0.5\linewidth]{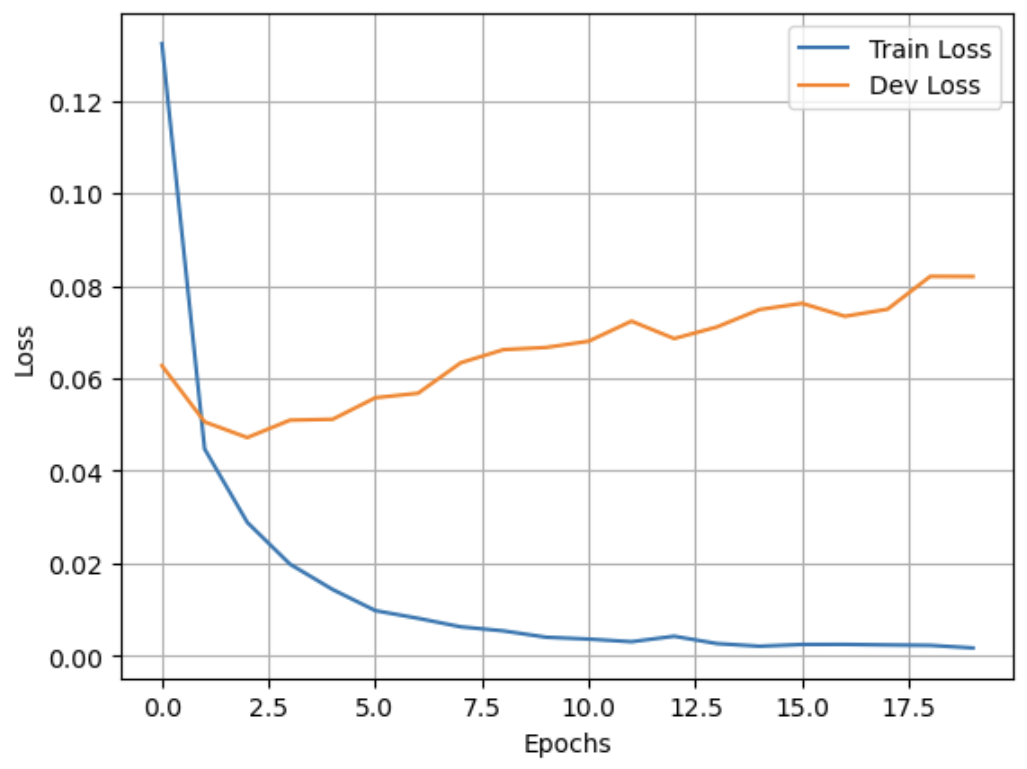} \hfill
\includegraphics[width=0.5\linewidth]{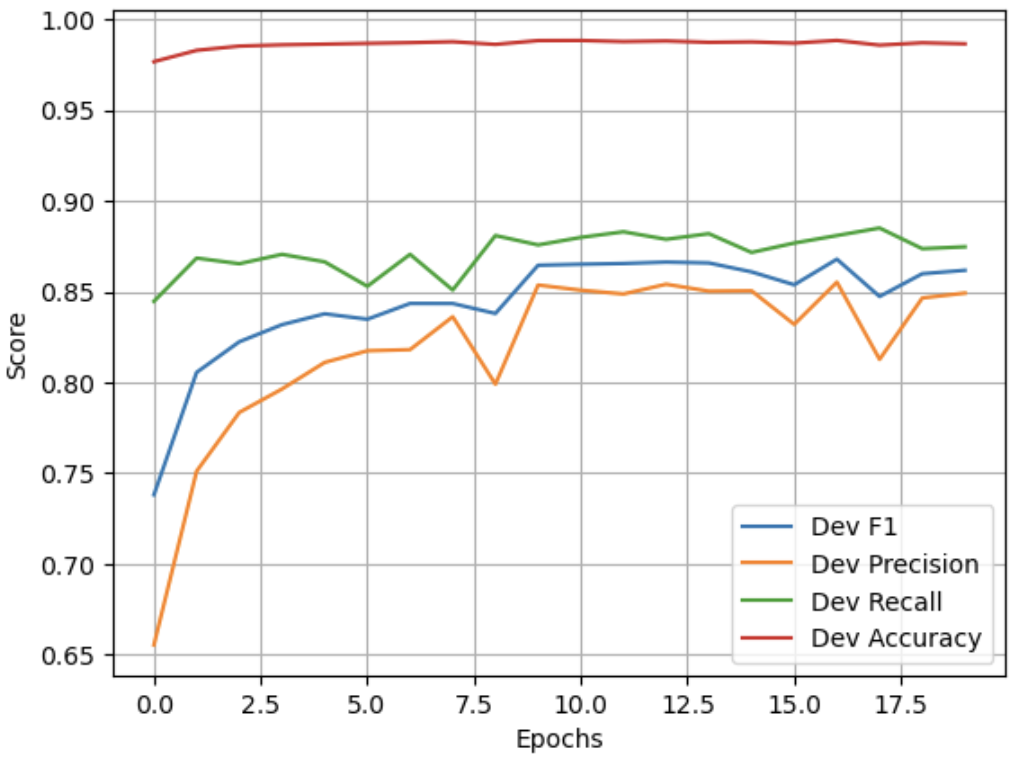}
\caption {Plots of training and validation loss (left) and validation metrics(right) over 20 epochs for the performance of BERT on the simple‐tag NER task.}
\label{tab:BERT_Simple}
\end{figure*}

\begin{figure*}[htbp]
  \includegraphics[width=0.5\linewidth]{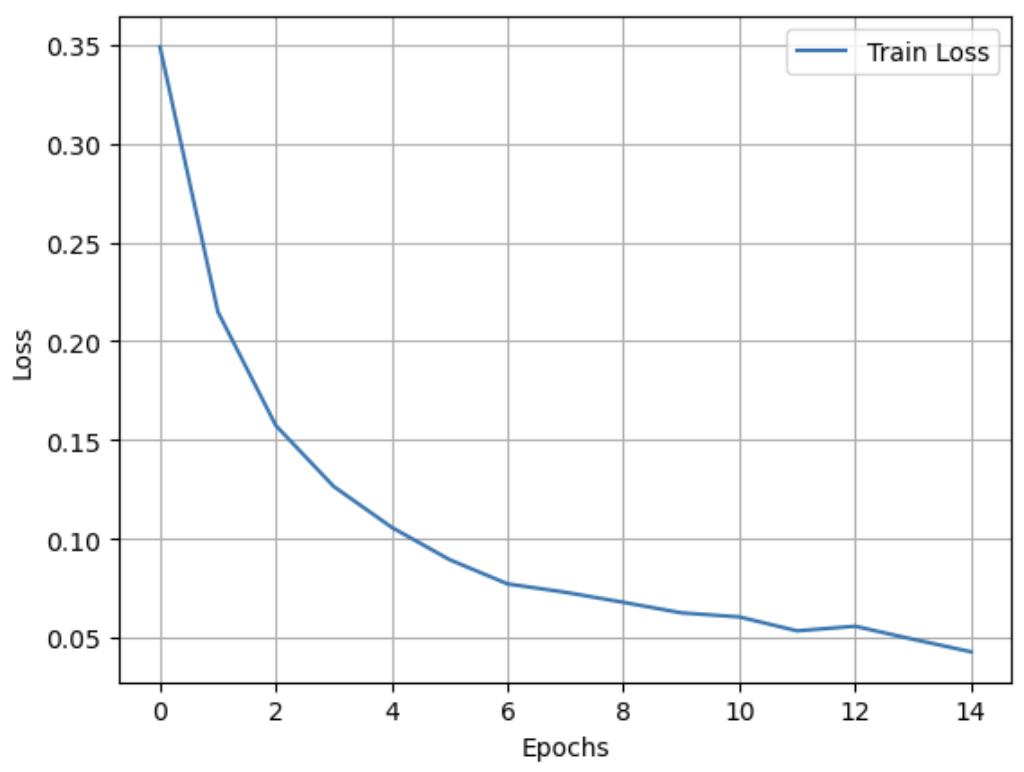} \hfill
  \includegraphics[width=0.5\linewidth]{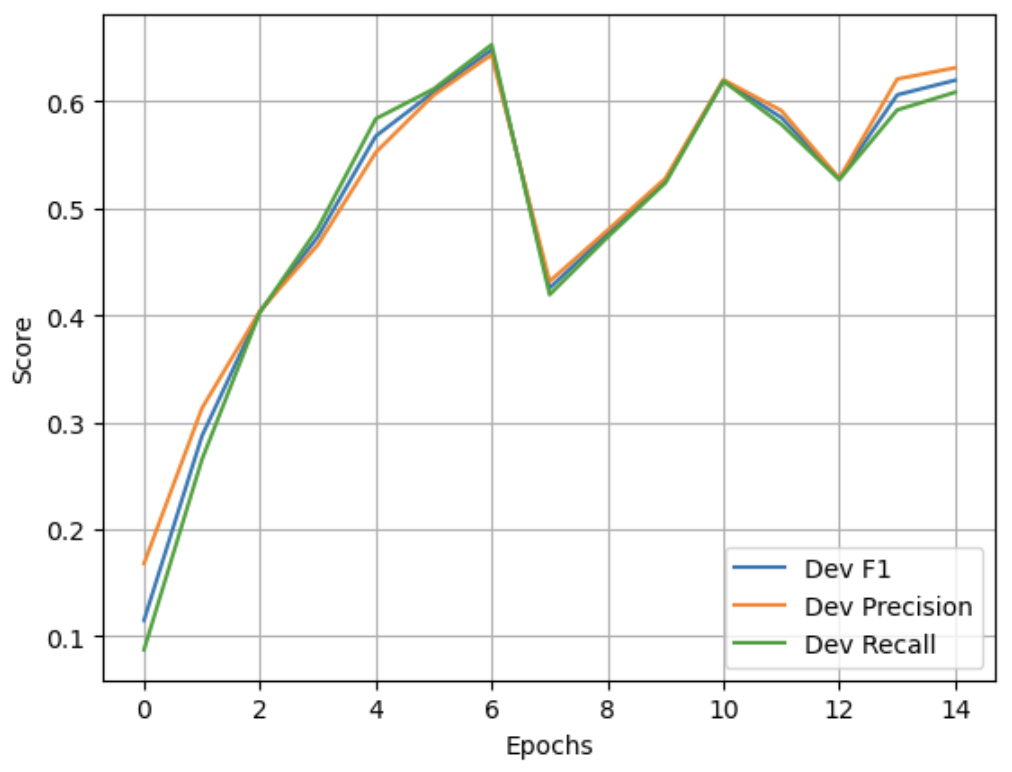}
  \caption {Plots of training loss (left) and validation metrics(right) over 15 epochs for the performance of T5 on full‐tag NER task. Early stopping at the seventh epoch.}
  \label{tab:T5_Full}
\end{figure*}

\begin{figure*}[htbp]
  \includegraphics[width=0.5\linewidth]{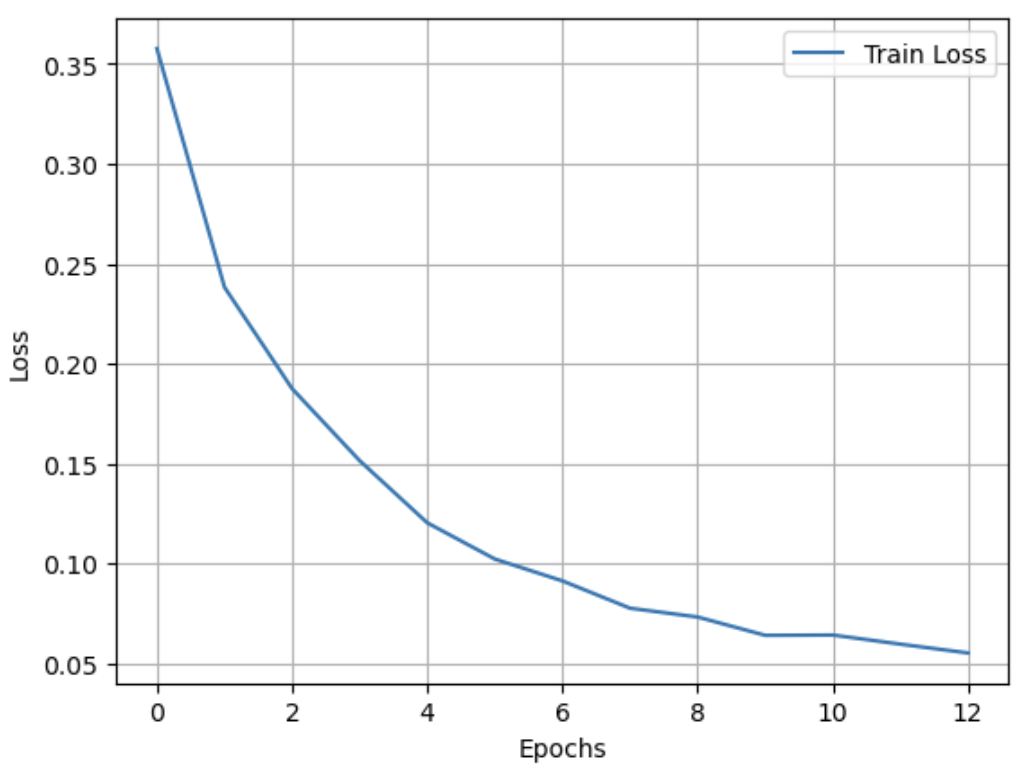} \hfill
  \includegraphics[width=0.5\linewidth]{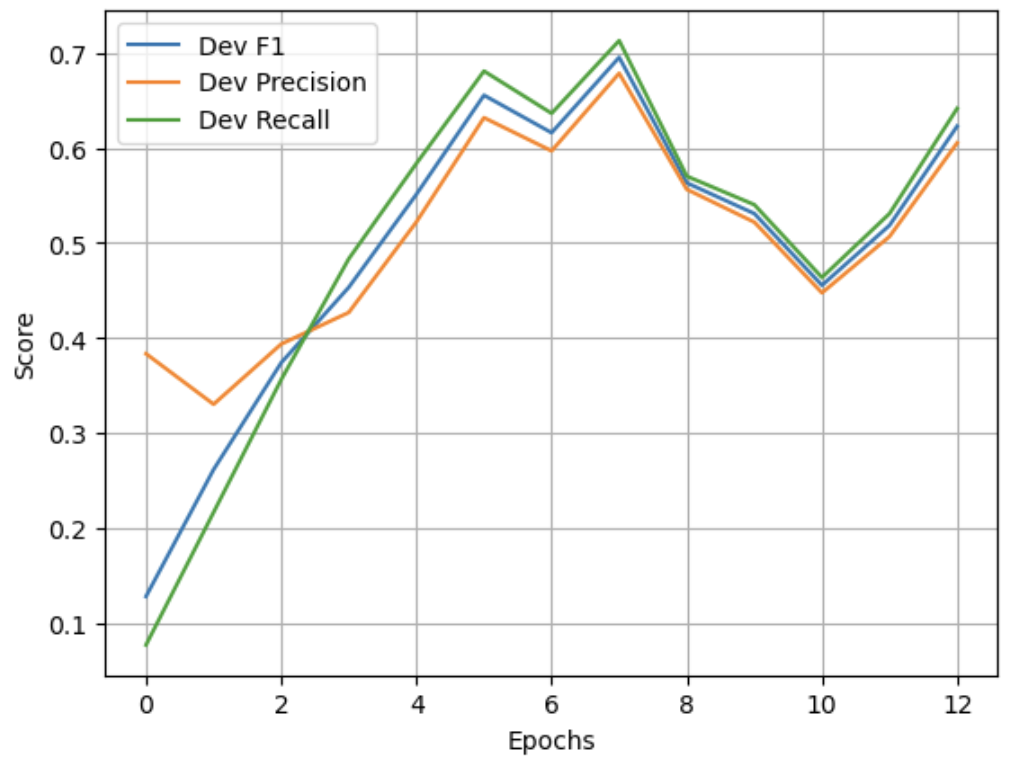}
  \caption {Plots of training loss (left) and validation metrics(right) over 13 epochs for the performance of T5 on simple‐tag NER task. Early stopping at the eighth epoch.}
\label{tab:T5_Simple}
\end{figure*}
\clearpage
\subsection{Error Examples}
\hspace{30pt}
\begin{figure*}[htbp]
\includegraphics[width=1\linewidth]{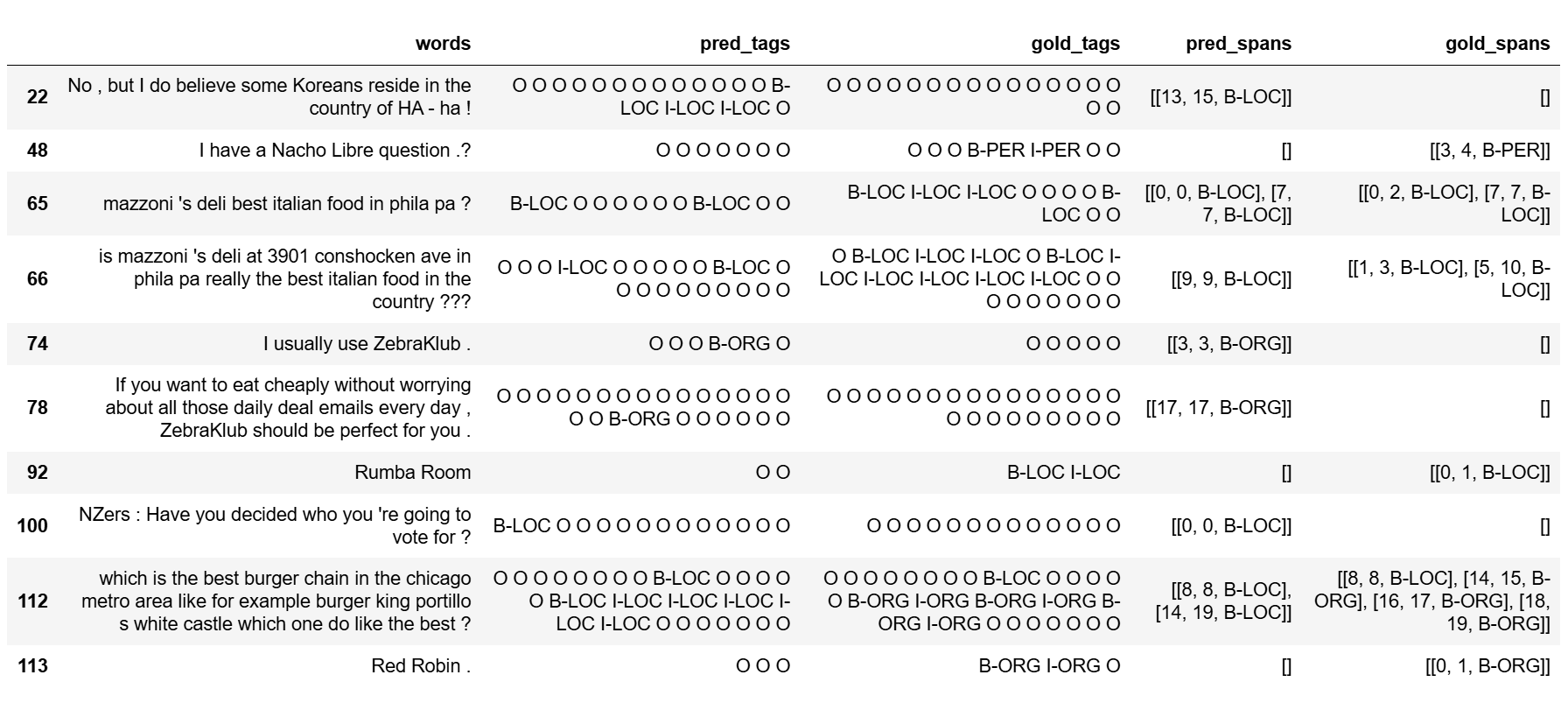} \hfill
\caption {Plot of top error examples of BERT in the Test dataset with full tags.}
\label{tag:error_Full_Test}
\hspace{30pt}
\end{figure*}
\begin{figure*}[htbp]
\includegraphics[width=1\linewidth]{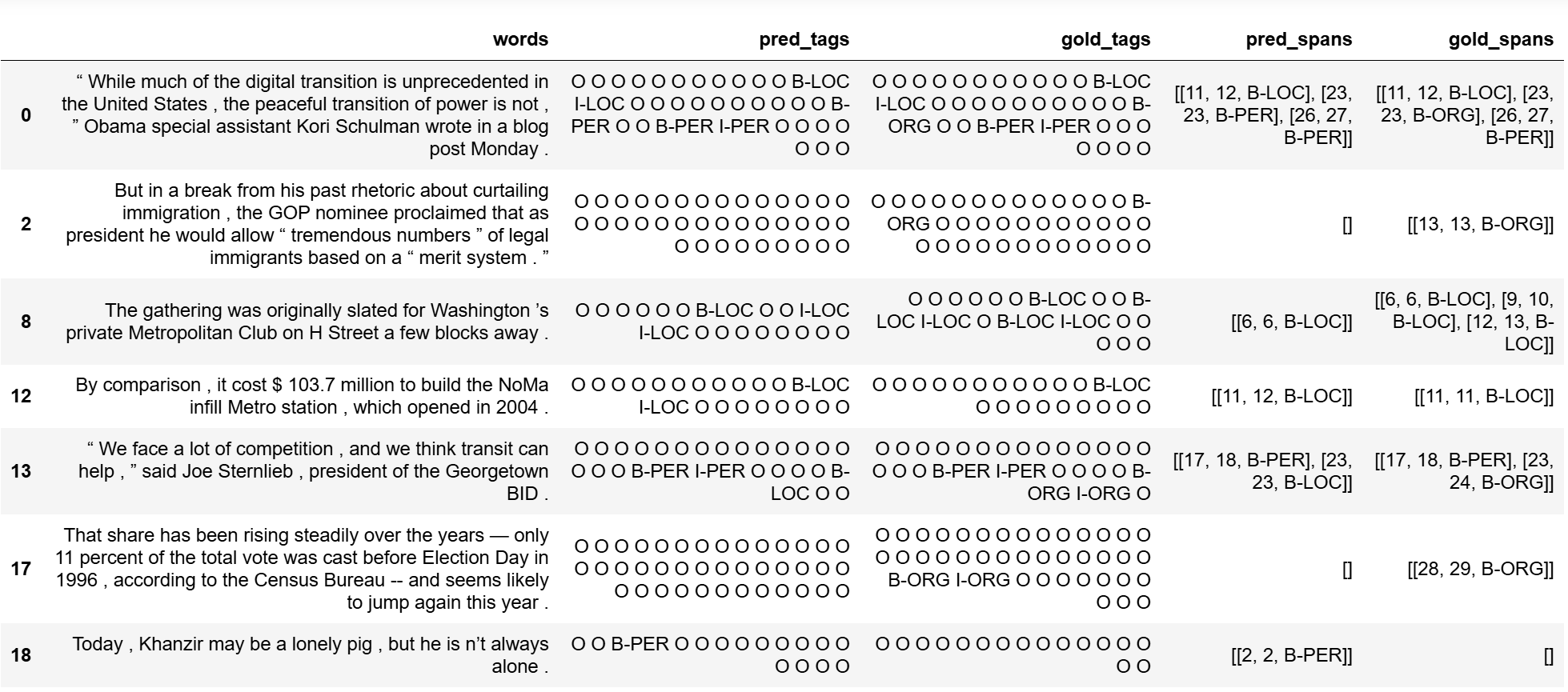} \hfill
\caption {Plot of top error examples of BERT in the OOD dataset with full tags.}
\label{tag:error_Full_OOD}
\end{figure*}
\begin{figure*}[htbp]
\includegraphics[width=1\linewidth]{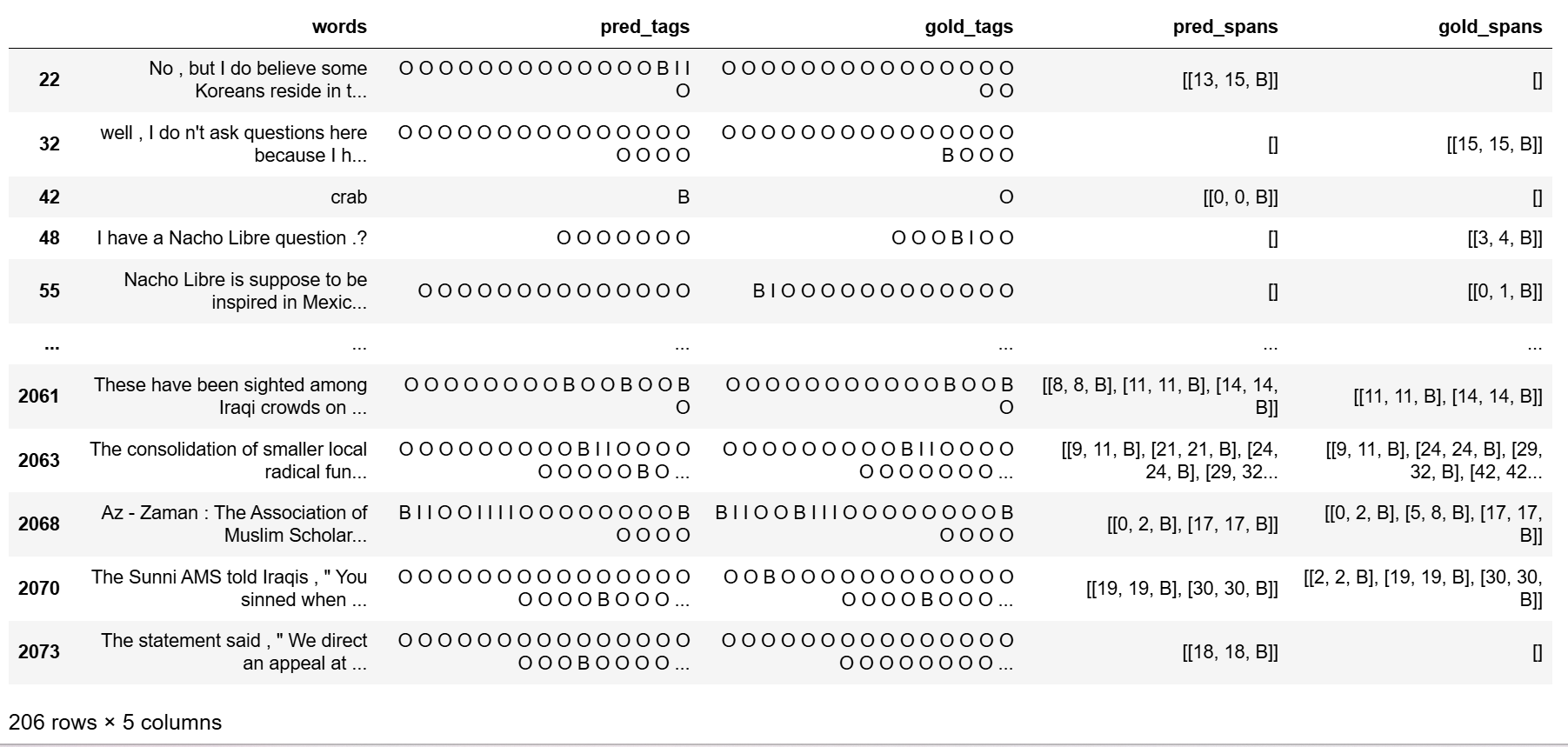} \hfill
\caption {Plot of error examples of BERT in the Test dataset with simple tags.}
\label{tag:error_Simple_Test}
\end{figure*}
\begin{figure*}[htbp]
\includegraphics[width=1\linewidth]{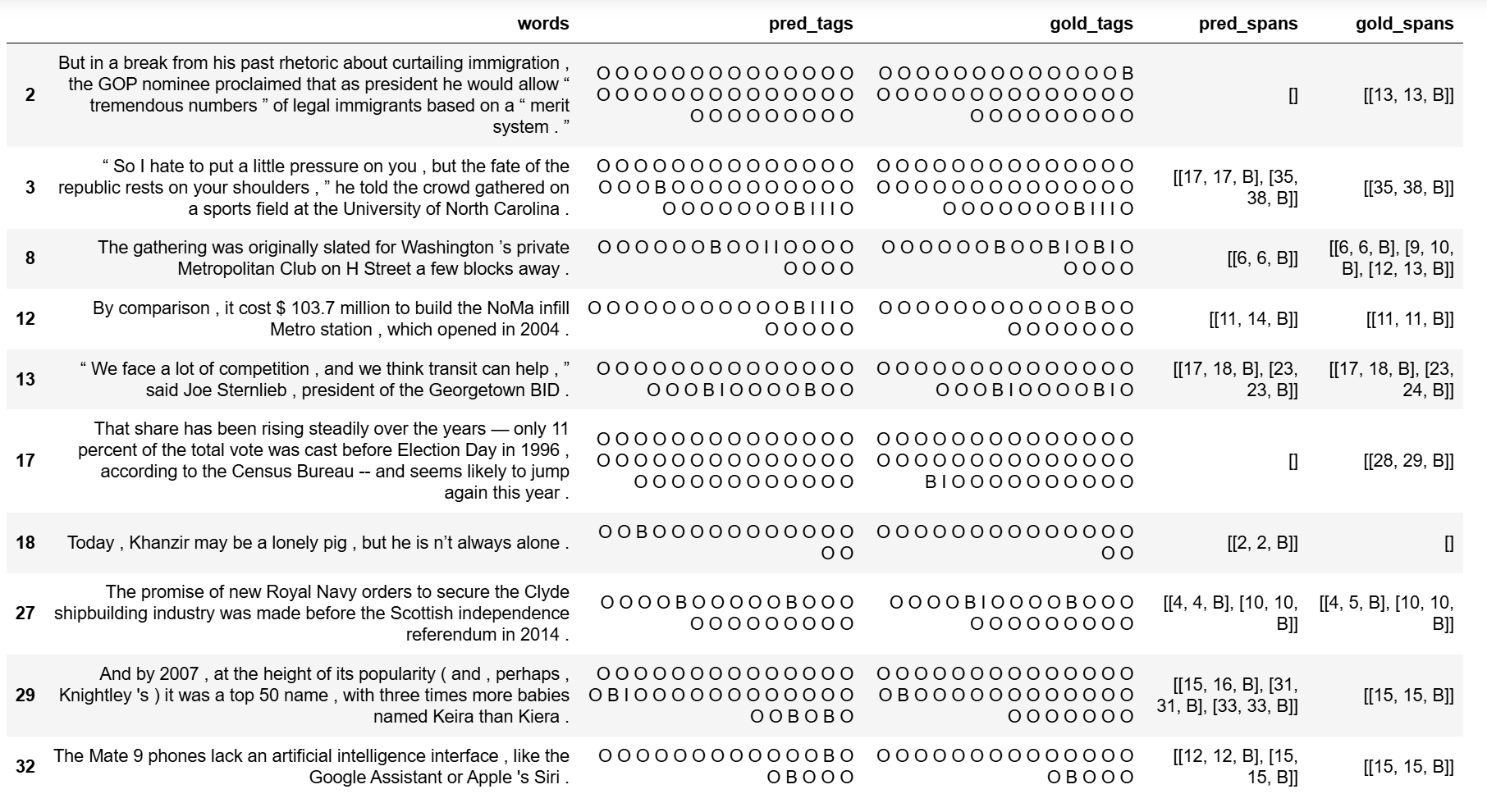} \hfill
\caption {Plot of top error examples of BERT in the OOD dataset with simple tags.}
\label{tag:error_Simple_OOD}
\end{figure*}

\end{document}